# Biomedical Mention Disambiguation using a Deep Learning Approach

Chih-Hsuan Wei
8600 Rockville Pike,
National Center for
Biotechnology Information
(NCBI) Bethesda,
Maryland, USA, 20894
Chih-Hsuan.Wei@nih.gov

Kyubum Lee
8600 Rockville Pike,
National Center for
Biotechnology Information
(NCBI) Bethesda,
Maryland, USA, 20894
Kyubum.Lee@nih.gov

Robert Leaman
8600 Rockville Pike,
National Center for
Biotechnology Information
(NCBI) Bethesda,
Maryland, USA, 20894
Robert.Leaman@nih.gov

Zhiyong Lu[*]
8600 Rockville Pike,
National Center for
Biotechnology Information
(NCBI) Bethesda,
Maryland, USA, 20894
Zhiyong.Lu@nih.gov

## ABSTRACT

Automatically locating named entities in natural language text – named entity recognition – is an important task in the biomedical domain. Many named entity mentions are ambiguous between several bioconcept types, however, causing text spans to be annotated as more than one type when simultaneously recognizing multiple entity types. The straightforward solution is a rule-based approach applying a priority order based on the precision of each entity tagger (from highest to lowest). While this method is straightforward and useful, imprecise disambiguation remains a significant source of error. We address this issue by generating a partially labeled corpus of ambiguous concept mentions. We first collect named entity mentions from multiple human-curated databases (e.g. CTDbase, gene2pubmed), then correlate them with the text mined span from PubTator to provide the context where the mention appears. Our corpus contains more than 3 million concept mentions that ambiguous between one or more concept types in PubTator ($\approx$ 3% of all mentions). We approached this task as a classification problem and developed a deep learning-based method which uses the semantics of the span being classified and the surrounding words to identify the most likely bioconcept type. More specifically, we develop a convolutional neural network (CNN) and along short-term memory (LSTM) network to respectively handle the semantic syntax features, then concatenate these within a fully connected layer for final classification. The priority ordering rule-based approach demonstrated F1-scores of 71.29% (micro-averaged) and 41.19% (macro-averaged), while the new disambiguation method demonstrated F1-scores of 91.94% (micro-averaged) and 85.42% (macro-averaged), a very substantial increase.

## KEYWORDS

Concept type disambiguation, natural language processing, convolutional neural network, recurrent neural network

## 1 Introduction

Mentions of concepts such as genes and diseases in the biomedical literature play a key role in knowledge integration and personalized medicine. Due to the rapid growth of the literature, automatically recognizing bioconcept mentions has become a very important task. The text mining tasks of named-entity recognition (NER) and normalization have been widely studied for a variety of bioconcept types, including genes/proteins [1-3], diseases [4, 5], chemicals [5-7], species [8, 9], sequence variations [10-13], and cell lines [14]. Most of these methods achieved over 80% of F1-score, a level of performance sufficient to allow the creation of several online systems [15-18] integrating the annotations from multiple NER taggers to support various downstream text mining tasks.

A variety of ambiguity issues accompany NER methods, making automated NER methods difficult. Among these issues are abbreviation ambiguity (e.g., "BD" can be Binswanger's disease and Behçet's disease) and term variants (e.g., erbb2 is also known as NEU; NGL; HER2; TKR1; CD340; HER-2; MLN 19; HER-2/neu). These have been explored well in previous work, such as Ab3P [19] for abbreviation ambiguity.

However, a specific term ambiguity issue that is rarely discussed is that many recognized mentions may be ambiguous among multiple bioconcept types. For instance, "CO2" is sometimes used as an abbreviation of the gene/protein "complement C2" (EntrezGene:717) but used for the chemical term "carbon dioxide" (MESH: D002245) in other articles. Previous work in PubTator [18], which integrates annotations from several NER taggers across articles in PubMed, used a straightforward rule-based approach for disambiguating bioconcept types. This approach is a priority order based on the precision of the NER taggers, ordered from the highest to the lowest. Normally, the tagger with the higher precision produces annotations with higher confidence. While helpful, many false positives and false negatives remain after using this approach. For instance, mutation annotations are prioritized ahead of all other bioconcepts, however we find false positives of mutations are often cell lines (e.g., A2780S in PMID: 25026335) or chemicals (e.g., C3368-A in PMID:7767952).

Unlike general named entity normalization studies which map the name entities found by NER to their corresponding concept identifiers [20-22], the task of biomedical concept disambiguation (BD) is to recognize the corresponding bioconcept type from a list of candidates (e.g., AP2 can be a name of gene (EntrezGene: 2167), chemical (MeSH:C523965) or cell-line). In other words, BD is a task for optimizing the performance of named-entity recognition and normalization. Most NER corpora are created using a small number of bioconcept types to limit the annotation work required, limiting their usefulness for BD. For example, the GENETAG corpus [23] annotates only genes, proteins, and the BC5CDR corpus [24] annotates only chemicals and diseases. Due to the lack of a comprehensive training corpus for BD, it is difficult to develop a model to recognize the corresponding bioconcept types of highly ambiguous mentions (e.g., abbreviations).

In response, we firstly generated a partially labeled corpus of ambiguous concept mentions by utilizing multiple human-curated databases and text mined spans of PubTator, and then applied convolutional neural networks (CNN) [25] and long short-term memory (LSTM) [26] to develop a classification method for this task. Our method analyses the semantic and syntactic logic of both the target mention and the words surrounding the target to identify the most likely bioconcept types for the mentions.

## 2 Benchmark Corpus

Table 1. The list of repositories we collected for building a comprehensive corpus

| Repository | Gene | Disease | Chemical | Species | Variant | Cell Line |
|---|---|---|---|---|---|---|
| gene2pubmed | √ | | | | | |
| GeneRIF [30] | √ | | | | | |
| gene_interactions [30] | √ | | | | | |
| CTDbase [29] | √ | √ | √ | | | |
| RGD [31] | √ | √ | | | √ | |
| BioGRID [32] | √ | | | | √ | |
| NCBI Taxonomy [33] | | | | √ | | |
| MeSH [34] | √ | √ | √ | √ | √ | √ |
| ZFIN [35] | √ | | | √ | | |
| ClinVar [36] | | | | | √ | |
| dbGAP [37] | | | | | √ | |
| dbSNP [38] | | | | | √ | |
| GWAS [39] | √ | √ | | | √ | |
| HPRD [40] | √ | √ | | | | |
| Cellosaurus [41] | | | | | | √ |

While the data used for method development and training is usually prepared manually, creating new manual annotation is highly labor intensive. In this work we instead collected the necessary biomedical named-entity data from multiple sources already annotated manually, such as MeSH [27], gene2pubmed [28] and CTDbase [29] as shown in Table 1. These repositories associate PubMed identifiers (PMIDs) with a concept or database identifier (accession ID), such as <PMID:10021333, GeneID:41066> from gene2pubmed (ftp://ftp.ncbi.nlm.nih.gov/gene/DATA/gene2pubmed.gz).

We have observed that the correct concept type for most ambiguous mentions can be identified by reviewing the context. However, none of the repositories record the location of the mentions corresponding to the concepts they annotate (i.e., offset and span). The absence of mention location significantly limits the immediate utility of the repository annotations for disambiguation. Unlike the manually annotated repositories, PubTator [18] provides the spans of the mentions which were automatically extracted by machine learning-based taggers (GNormPlus [3] for genes, tmVar [12, 13] for variants, SR4GN [9] for species, DNorm [4] for diseases and tmChem [6] for chemicals). Cell lines are another common concept type that are frequently ambiguous. In this work, we rebuilt an NER model in TaggerOne [5] to recognize cell line mentions. The model is trained and evaluated by the released corpus of BioCreative Bio-ID task [42] and obtained 83.10% of F1-score. These taggers were previously evaluated and achieved 80-90% of F1-scores in normalization results. To obtain the spans of the concepts in the repositories, we utilized the spans recorded in PubTator. For example, while MeSH associates PMID:23262785 with "Breast Neoplasms" (MeSH:D001943), the mention recognized by DNorm for MeSH:D001943 is "breast cancer." We therefore consider the span of the mention "breast cancer" as the span for MeSH ID:D001943 in PMID:23262785 (as shown in Figure 1).

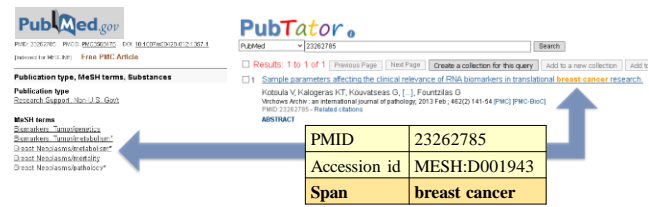

Figure 1. An example of the confirmed MESH annotation with span

Figure 2 shows the number of annotations in all the collected repositories and in PubTator respectively. The overlapping area represents the repository annotations that can be paired with spans in PubTator. For example, <PMID:10022874, NCBIGene:1977, Spans: "Eukaryotic translation initiation factor 4E"&"eIF4E">. Correlating annotations from the repositories and PubTator resulted in nearly 24 million repository records (25.6%) from 13 million articles being associated with spans from PubTator.

We further separated the records into individual spans. For example, <PMID:10022874, NCBIGene:1977, Spans: "Eukaryotic translation initiation factor 4E"&"eIF4E"> can be separated into two individual records <PMID:10022874, NCBIGene:1977, Spans: "Eukaryotic translation initiation factor 4E"> and <PMID:10022874, NCBIGene:1977, Spans:"eIF4E">. We obtained a total of 33,173,360 records with individual spans. To focus on the subset of records representing ambiguous annotations, we filtered the spans to only retain spans tagged with multiple entity types, such as "XPID" in PMID: 23378296, which is recognized as both a disease and a gene in PubTator ("XPID" is manually annotated as a gene in MeSH). After filtering, 219,247 annotations remain.

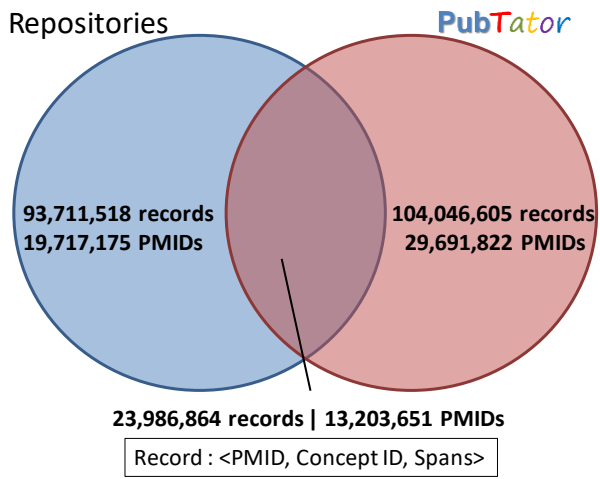

Figure 2. The annotations in repositories and the annotations in PubTator (with spans).

Table 2. The number of annotations <PMID, Concept ID, Spans> in repositories.

| Bioconcept | # of annotations with spans from multiple repositories | Individual mentions with spans found in PubTator | Ambiguous mentions with spans |
|---|---|---|---|
| Gene | 2,076,650 | 3,093,005 | **79,418** |
| Disease | 9,683,680 | 14,356,863 | **27,591** |
| Species | 4,677,865 | 6,575,790 | **65,113** |
| Chemical | 7,431,195 | 9,014,513 | **46,400** |
| Variation | 92,411 | 108,019 | **262** |
| Cell line | 20,76,650 | 25,170 | **463** |
| Total | 23,986,864 | 33,173,360 | **219,247** |
| Articles in Total | | 13,203,651 | **184,671** |

Table 3. The number of annotations in the training and test sets, for each concept type.

| Bioconcept | Random sampling | | Independent sampling | |
|---|---|---|---|---|
| | Training set | Test set | Training set | Test set |
| Gene | 63,521 | 20,952 | 58,466 | 15,897 |
| Disease | 22,065 | 2,158 | 25,433 | 5,526 |
| Species | 52,175 | 12,479 | 52,634 | 12,938 |
| Chemical | 37,041 | 10,120 | 36,280 | 9,359 |
| Variation | 210 | 62 | 200 | 52 |
| Cell line | 385 | 74 | 389 | 78 |
| Total | 175,397 | 45,845 | 173,402 | 43,850 |

We created two versions of the training and test sets, using different sampling strategies, as shown in Table 3. We first selected 20% of the ambiguous mentions with spans for testing and the other 80% for training/validation (random sampling). However, under random sampling most of the mentions in the test set would also appear in the training set if the test mentions. To sufficiently reflect real-world performance, we prepared another training/test set split (independent sampling), where the test set only contains mentions that do not appear in the training set, though they were selected randomly otherwise.

## 3 Method

As illustrated in Figure 3, we first collected the manual annotations from various repositories (in Figure 3a) and associated them with the text mined mention spans from PubTator (in Figure 3b), and further separated the corpus into training and test sets (in Figure 3c). We applied the training set for the development of the classifier and the test set for performance evaluation (in Figure 3d).

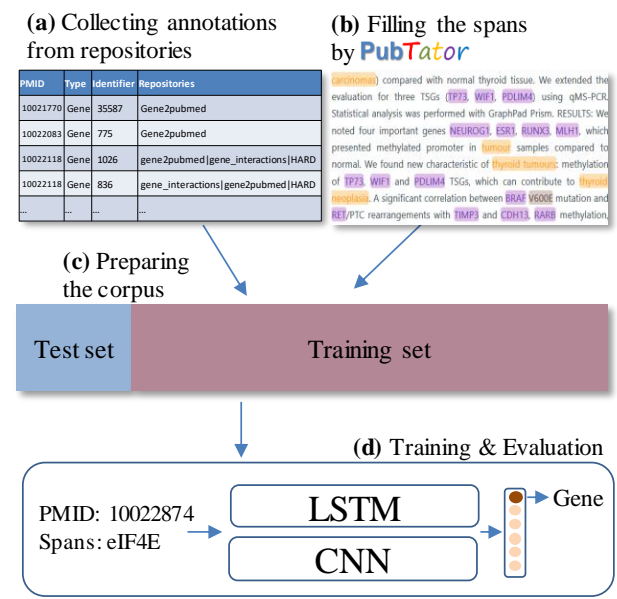

Figure 3. Method overview.

As shown in Figure 4, we used the surrounding words before and after the target mention within the context window (size = 10). For an example in the first sentence of PMID:10022874 "*Eukaryotic translation initiation factor 4E (***eIF4E***) binds to the mRNA 5' cap and brings the mRNA into a complex with other protein synthesis initiation factors and ribosomes.*", the context words before the target mention "eIF4E", are "Eukaryotic translation initiation factor 4E eIF4E" and the context words after it are "eIF4E binds to the mRNA 5 cap and brings the RNA". Generally, LSTM is more effective than CNN on sequential input, because LSTMs model sequential data directly by considering the input features of the previous token, while CNNs only consider the input features of the current token. Thus, we used LSTM for the surrounding words features and CNN for semantic features. To emphasize the importance of the target mention, we applied two LSTM models with opposite directions to the forward and backward strings. The last states of the two LSTM models are passed to the final fully connected layer for predicting the concept type of the mention.

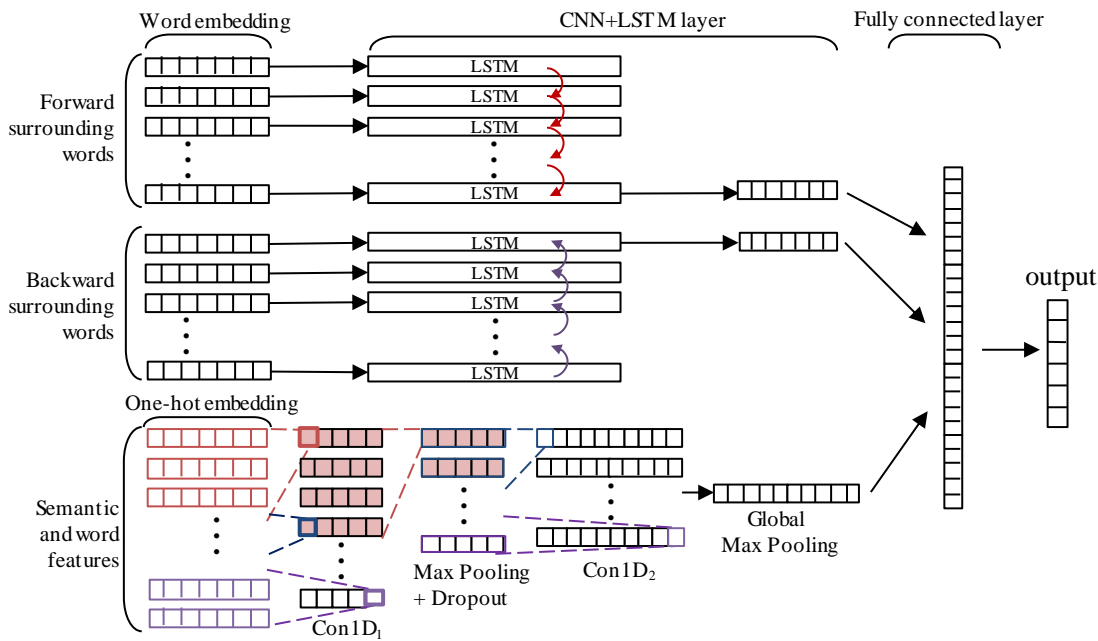

Figure 4. The system overview that includes three major layers: word embedding layer, CNN+LSTM layer and concatenation layer.

Table 4. The list of the semantic and word features

| Feature types | Features |
|---|---|
| Semantic features | Concept types detected by NER taggers |
| | Concept identifiers detected by NER taggers |
| | The concept type of the full name detected by NER taggers |
| | The concept identifier of the full name detected by NER taggers |
| Word features | Prefix (Length = 1~3) |
| | Suffix (Length = 1~3) |

We further applied a CNN model on the semantic and word features collected by the rules list in Table 4. Word features includes the prefixes and the suffixes of all tokens of the target mention. Semantic features include those generated by the NER taggers, including concept type, identifiers of the target mentions and its correspond full name recognized by Ab3P [19]. Our method uses word embeddings to translate word to vectors, which generalizes well for biomedical text. The embedding layer of the word and semantic features use one-hot embeddings.

Due to the lack of the variant and cell line records, two types are weakened significantly. To increase the number of training records of the two types, we repeatedly loaded the records of the two types in training sets for 10 times. We implemented the CNN/LSTM model using Keras and the TensorFlow library. Table 5 summarizes the hyperparameters we applied in our network. Our word embedding was the 200-dimensional vector which used the word2vec [43] skip-gram implementation on all PubMed abstracts and Wikipedia pages. We also applied dropout to reduce the effect of overfitting.

Table 5. The parameters of the CNN model

| Hyper-parameter | | Value |
|---|---|---|
| Embedding dimension | Word embedding | 200 |
| | Ono-hot embedding | 200 |
| Convolutional layer 1 (Con1D$_1$) | Number of filters | 200 |
| | Number of kernels | 5 |
| | Max pooling size | 5 |
| | Dropout rate | 0.2 |
| Convolutional layer 2 (Con1D$_2$) | Number of filters | 1000 |
| | Number of kernels | 5 |
| | Global Max pooling | |
| LSTM layer | Number of units | 128 |
| | Dropout rate | 0.2 |
| | Recurrent dropout | 0.2 |
| Concatenate layer | Number of units | 1256 |
| | Activation | Relu |
| Fully connected layer | Number of units | 5 |
| | Activation | Softmax |

### 3.1 Baseline methods

We applied three additional methods for comparison: a rule-based method used in the original version of PubTator, maximum entropy classification (MaxEnt) [44] and BioBERT [45]. The rule-based method is a priority order based on the precision of each concept tagger, which is mutation > species > gene > chemical > disease > cell line. Secondly, we applied MaxEnt and to handle this task as a multiple classification problem. The MaxEnt has been widely used to deal with multi-type text classification problems in past decades and obtained comparable performance. We also applied BioBERT, which is a BERT model retrained on the entire PubMed/PMC and has obtained significant improvements on named-entity recognition, relation extraction, and question answering tasks. We converted the task to a sentence classification problem for BERT and all the contextual and semantic features are arranged as a sentence input. We also built a model which using a CNN layer for surrounding words and semantic features to determine how the LSTM helps the performance.

## 4 Result

### 4.1 Evaluation metrics

We used the F1-score to evaluate the performance of the methods. The precision (P) represents the percentage of records correctly predicted divided by the total number of predictions. The recall (R) represents the percentage of records correctly predicted divided by the total number of records. Given the precision and recall, F1-score can be calculated as $(P \times R \times 2) \div (P+R)$. Given the lower number of variant and cell line type mentions in the test set, we also calculated the macro-averaged P, R, and F1-score, to give equal weight to each type. Macro-averaging is to average the P, R, and F1-score in all concept types.

### 4.2 Evaluation

Table 6. Performance on the benchmark corpus

|  | Random sampling | | Independent sampling | |
| --- | --- | --- | --- | --- |
|  | Micro F1-score | Macro F1-score | Micro F1-score | Macro F1-score |
| CNN+LSTM | **93.56%** | **92.55%** | **91.94%** | **85.42%** |
| CNN only | 93.34% | 91.09% | 91.78% | 84.88% |
| BioBERT | 91.36% | 88.26% | 89.31% | 82.22% |
| MaxEnt | 92.38% | 87.69% | 89.03% | 82.44% |
| Rule-based | 68.99% | 41.34% | 71.29% | 41.19% |

In our experiments, the performance of all the machine learning-based (ML) methods on the randomly selected set provided excellent performance – F1-scores of over 90% – in both micro and macro averages. The performance dropped slightly, by 2% micro F1-score and 7% of macro F1-score, on the independent sampling set, indicating that the ML methods are robust in handling the unknown mentions. In addition, using the LSTM layer for surrounding words features slightly outperforms the model using CNN layers for all the features. Overall, our method presents the highest performance in both the random and independent sampling sets and can significantly improve the quality of PubTator ambiguous annotations (~25% improvement on micro F1-score and 40-50% improvement on macro F1-score).

Furthermore, we separated the concept types to see the individual performance. Four concepts (i.e., species, mutation, chemical and gene) present higher performance than the overall macro-averaged F1-score, but disease and cell line types both present lower performances. The reason of the lower recall of cell line may due to the insufficient training data for the cell line type, which is less than 1% of the other types. The majority of the errors for cell line mentions are incorrectly predicting genes or chemicals, because chemicals and genes in the training set are frequently recognized by cell line tagger incorrectly. Disease scored the lowest precision, because many of the human disease names are confused with disease-causing viruses or bacteria (species names). More discussions of these points are in the error analysis.

Table 7. Performances of individual concepts of the CNN+LSTM method in independent test set

| Concept type | Precision | Recall | F1-score |
| --- | --- | --- | --- |
| Disease | 71.37% | 71.21% | 71.29% |
| Species | 99.60% | 94.06% | 96.75% |
| Mutation | 93.10% | 87.10% | 90.00% |
| Cell line | 93.75% | 60.81% | 73.77% |
| Chemical | 92.79% | 83.85% | 88.10% |
| Gene | 88.77% | 96.74% | 92.59% |
| Micro average | 91.93% | 91.94% | 91.94% |
| Macro average | 89.90% | 82.30% | 85.42% |

### 4.3 Error analysis

Table 8. Bioconcept disambiguation error types in test set

| Description | # | % |
| --- | --- | --- |
| Conflict concepts are correct | 143 | 72% |
| One concept is a substring and associate with the other | 31 | 16% |
| Abbreviation is not the same concept type to the full name | 15 | 8% |
| Others | 11 | 3% |
| Total reviewed | 200 | 100% |

To understand the causes of errors made by our method, we manually reviewed the incorrectly classified examples from the independent set made by the model which presented the best micro F1-score (91.93%), and further classified 200 randomly selected errors into several categories. In our observation, the most frequent overlapping types are species and disease. For example, "African Horse Sickness virus" is a name of species, but it can be also the name of an infectious disease. However, the repositories only one of the types were annotated in the repositories (i.e. either species or disease). In the other case, one concept is a substring of the other concept and is strongly associated with it, such as the gene name "breast cancer 2", which includes the disease name "breast cancer" as a substring. The other frequent errors are due to conflicts between the abbreviation and its full name. For example, "Alkaptonuria" is a disease, but the abbreviation AKU is also associated with the gene that causes

Alkaptonuria. In such cases, the classifiers are very easily confused.

## 5   Conclusion

Bioconcept disambiguation is an important task in the field of biomedical text mining. Our previous bioconcept tagging work found more than 3 million ambiguous concept mentions in PubMed and 17 million ambiguous concept mentions in PMC full text. In this study, we presented a deep learning-based method to address the task of bioconcept disambiguation, applying CNNs and LSTMs to respectively capture the semantic and syntactic features. To reduce the effort required to manually curate a benchmark set, we proposed a method to generate partially labeled data by gathering manual annotations from multiple existing repositories, then integrating them with the spans of PubTator annotations. The rule-based priority ordering approach demonstrated an F1-score of 71.29% on this dataset, while our proposed disambiguation method demonstrated an F1-score of 91.94%, a very substantial improvement.

There are two primary use cases for this work. One is to optimize the performance on ambiguous annotations when using several taggers to identify multiple concept types. For example, this method significantly improves the NER performance of our recently published text mining system PubTator Central [46]. The other primary use case is as a post-processing tool for a single NER tagger (e.g., TaggerOne), to improve the performance on the desired entity types by filtering predicted mentions more likely to be another type (false positives).

An obvious limitation of this work is the relative lack of the ambiguous spans that could be collected form the existing repositories for cell lines (only 463 annotations available) and variants (only 262 annotations available). In the future, we intend to increase the number of annotations available for cell lines and variants through manual curation.

## ACKNOWLEDGMENTS

Funding: This research was supported by the NIH Intramural Research Program, National Library of Medicine.

## REFERENCES

[1] J. Hakenberg et al., "The GNAT library for local and remote gene mention normalization," Bioinformatics, vol. 27, no. 19, pp. 2769-2771, 2011.
[2] R. Leaman and G. Gonzalez, "BANNER: an executable survey of advances in biomedical named entity recognition," in Biocomputing 2008: World Scientific, 2008, pp. 652-663.
[3] C.-H. Wei, H.-Y. Kao, and Z. Lu, "GNormPlus: An Integrative Approach for Tagging Genes, Gene Families, and Protein Domains," BioMed Research International, no. Text Mining for Translational Bioinformatics special issue, p. 918710, 2015.
[4] R. Leaman, R. I. Doğan, and Z. Lu, "DNorm: disease name normalization with pairwise learning to rank," Bioinformatics, vol. 29, no. 22, pp. 2909-2917, 2013.
[5] R. Leaman and Z. Lu, "TaggerOne: joint named entity recognition and normalization with semi-Markov Model," Bioinformatics, vol. 32, no. 18, pp. 2839-2846, 2016.
[6] R. Leaman, C.-H. Wei, and Z. Lu, "tmChem: a high performance approach for chemical named entity recognition and normalization," Journal of cheminformatics, vol. 7, no. supplement 1, 2015.
[7] T. Rocktäschel, M. Weidlich, and U. Leser, "ChemSpot: a hybrid system for chemical named entity recognition," Bioinformatics, vol. 28, no. 12, pp. 1633-1640, 2012.
[8] M. Gerner, G. Nenadic, and C. M. Bergman, "LINNAEUS: A species name identification system for biomedical literature," BMC Bioinformatics, vol. 11, p. 85, 2010.
[9] C.-H. Wei, H.-Y. Kao, and Z. Lu, "SR4GN: a species recognition software tool for gene normalization," PloS one, vol. 7, no. 6, p. e38460 2012.
[10] J. M. Cejuela et al., "nala: text mining natural language mutation mentions," Bioinformatics, vol. 33, no. 12, pp. 1852-1858, 2017.
[11] P. Thomas, T. Rocktäschel, Y. Mayer, and U. Leser, "SETH: SNP extraction tool for human variations," ed, 2014.
[12] C.-H. Wei, B. R. Harris, H.-Y. Kao, and Z. Lu, "tmVar: A text mining approach for extracting sequence variants in biomedical literature," Bioinformatics, vol. 29, no. 11, pp. 1433-1439, 2013.
[13] C.-H. Wei, L. Phan, J. Feltz, R. Maiti, T. Hefferon, and Z. Lu, "tmVar 2.0: integrating genomic variant information from literature with dbSNP and ClinVar for precision medicine," Bioinformatics, vol. 34, no. 1, pp. 80-87, 2017.
[14] H. Stachelscheid et al., "CellFinder: a cell data repository," Nucleic acids research, vol. 42, no. D1, pp. D950-D958, 2014.
[15] D. Campos, J. Lourenço, S. Matos, and J. L. Oliveira, "Egas: a collaborative and interactive document curation platform," Database, vol. 2014, 2014.
[16] F. Rinaldi, G. Schneider, K. Kaljurand, S. Clematide, T. Vachon, and M. Romacker, "Ontogene in biocreative ii. 5," IEEE/ACM Transactions on Computational Biology and Bioinformatics (TCBB), vol. 7, no. 3, pp. 472-480, 2010.
[17] P. Thomas, J. Starlinger, A. Vowinkel, S. Arzt, and U. Leser, "GeneView: a comprehensive semantic search engine for PubMed," Nucleic acids research, vol. 40, no. W1, pp. W585-W591, 2012.
[18] C.-H. Wei, H.-Y. Kao, and Z. Lu, "PubTator: a Web-based text mining tool for assisting Biocuration," Nucleic acids research, vol. 41, no. W1, pp. W518-W522, 2013.
[19] S. Sohn, D. C. Comeau, W. Kim, and W. J. Wilbur, "Abbreviation definition identification based on automatic precision estimates," BMC bioinformatics, vol. 9, no. 1, p. 402, 2008.
[20] C. Arighi et al., "Bio-ID track overview," annotation, vol. 501, p. 1, 2017.
[21] C.-H. Wei et al., "Overview of the BioCreative V chemical disease relation (CDR) task," in Proceedings of the fifth BioCreative challenge evaluation workshop, 2015, pp. 154-166.
[22] Z. Lu et al., "The gene normalization task in BioCreative III," BMC bioinformatics, vol. 12, no. 8, p. S2, 2011.
[23] L. Tanabe, N. Xie, L. H. Thom, W. Matten, and W. J. Wilbur, "GENETAG: a tagged corpus for gene/protein named entity recognition," BMC bioinformatics, vol. 6, no. 1, p. S3, 2005.
[24] J. Li et al., "BioCreative V CDR task corpus: a resource for chemical disease relation extraction," Database, vol. 2016, 2016.
[25] Y. Kim, "Convolutional neural networks for sentence classification," arXiv preprint arXiv:1408.5882, 2014.
[26] S. Hochreiter and J. Schmidhuber, "Long short-term memory," Neural computation, vol. 9, no. 8, pp. 1735-1780, 1997.
[27] C. E. Lipscomb, "Medical subject headings (MeSH)," Bulletin of the Medical Library Association, vol. 88, no. 3, p. 265, 2000.
[28] N. R. Coordinators, "Database resources of the national center for biotechnology information," Nucleic acids research, vol. 46, no. Database issue, p. D8, 2018.
[29] A. Davis et al., "CTD-Comparative Toxicogenomics Database," ed.
[30] G. R. Brown et al., "Gene: a gene-centered information resource at NCBI," Nucleic acids research, vol. 43, no. D1, pp. D36-D42, 2014.
[31] R. Nigam et al., "Rat Genome Database: a unique resource for rat, human, and mouse quantitative trait locus data," Physiological genomics, vol. 45, no. 18, pp. 809-816, 2013.
[32] A. Chatr-Aryamontri et al., "The BioGRID interaction database: 2017 update," Nucleic acids research, vol. 45, no. D1, pp. D369-D379, 2017.
[33] S. Federhen, "The NCBI taxonomy database," Nucleic acids research, vol. 40, no. D1, pp. D136-D143, 2011.
[34] I. K. Dhammi and S. Kumar, "Medical subject headings (MeSH) terms," Indian journal of orthopaedics, vol. 48, no. 5, p. 443, 2014.
[35] L. Ruzicka et al., "ZFIN, The zebrafish model organism database: Updates and new directions," genesis, vol. 53, no. 8, pp. 498-509, 2015.
[36] M. J. Landrum et al., "ClinVar: improving access to variant interpretations and supporting evidence," Nucleic acids research, vol. 46, no. D1, pp. D1062-D1067, 2017.
[37] K. M. Wong et al., "The dbGaP data browser: a new tool for browsing dbGaP controlled-access genomic data," Nucleic acids research, vol. 45, no. D1, pp. D819-D826, 2016.

[38] S. T. Sherry et al., "dbSNP: the NCBI database of genetic variation," Nucleic acids research, vol. 29, no. 1, pp. 308-311, 2001.
[39] J. MacArthur et al., "The new NHGRI-EBI Catalog of published genome-wide association studies (GWAS Catalog)," Nucleic acids research, vol. 45, no. D1, pp. D896-D901, 2016.
[40] T. Keshava Prasad et al., "Human protein reference database—2009 update," Nucleic acids research, vol. 37, no. suppl_1, pp. D767-D772, 2008.
[41] A. Bairoch, "The cellosaurus, a cell-line knowledge resource," Journal of biomolecular techniques: JBT, vol. 29, no. 2, p. 25, 2018.
[42] C. Arighi et al., "Bio-ID track overview," Cell, vol. 482, no. 7310, p. 376, 2017.
[43] T. Mikolov, I. Sutskever, K. Chen, G. S. Corrado, and J. Dean, "Distributed representations of words and phrases and their compositionality," in Advances in neural information processing systems, 2013, pp. 3111-3119.
[44] K. Nigam, J. Lafferty, and A. McCallum, "Using maximum entropy for text classification," in IJCAI-99 workshop on machine learning for information filtering, 1999, vol. 1, pp. 61-67.
[45] J. Lee et al., "BioBERT: pre-trained biomedical language representation model for biomedical text mining," arXiv preprint arXiv:1901.08746, 2019.
[46] C.-H. Wei, A. Allot, R. Leaman, and Z. Lu, "PubTator Central: Automated Concept Annotation for Biomedical Full Text Articles," Nucleic acids research, no. web server issue, p. in press, 2019.